\documentclass[conference]{IEEEtran}

\IEEEoverridecommandlockouts
\usepackage{cite}
\usepackage{amsmath,amssymb,amsfonts}
\usepackage{graphicx}
\usepackage{textcomp}
\usepackage{xcolor}

\usepackage{url}

\usepackage{amssymb}
\usepackage{amsmath}
\usepackage{amsthm}
\usepackage{float}
\usepackage{amsfonts}

\usepackage{latexsym}
\usepackage{amssymb}
\usepackage{amsmath}
\usepackage{amsthm}
\usepackage{booktabs}
\usepackage{enumitem}
\usepackage{graphicx}
\usepackage{color}
\usepackage{bbm}
\usepackage{comment}

\usepackage{algorithm}%
\usepackage{algorithmicx}%

\usepackage[noend]{algpseudocode}
\algnewcommand{\LineComment}[1]{\Statex \(\triangleright\) #1}
\algnewcommand\algorithmicinput{\textbf{Input:}}
\algnewcommand\algorithmicoutput{\textbf{Output:}}
\algnewcommand\Input{\item[\algorithmicinput]}
\algnewcommand\Output{\item[\algorithmicoutput]}

\def\algbackskip{\hskip-\ALG@thistlm}

\def\BibTeX{{\rm B\kern-.05em{\sc i\kern-.025em b}\kern-.08em
    T\kern-.1667em\lower.7ex\hbox{E}\kern-.125emX}}
\begin{document}

\title{Interpretable and Accurate Hybrid Decision Trees with Selective Case-Based Splits}
\author{\IEEEauthorblockN{Cristiano Landi}
\IEEEauthorblockA{\textit{University of Pisa and ISTI-CNR } \\
Pisa, Italy \\
cristiano.landi@phd.unipi.it}
\and
\IEEEauthorblockN{Alessio Cascione}
\IEEEauthorblockA{\textit{University of Pisa} \\
Pisa, Italy \\
alessio.cascione@phd.unipi.it}
\and
\IEEEauthorblockN{Riccardo Guidotti}
\IEEEauthorblockA{\textit{University of Pisa and ISTI-CNR } \\
Pisa, Italy \\
riccardo.guidotti@unipi.it}
}

\maketitle
\begingroup
\renewcommand{\thefootnote}{}
\footnotetext[0]{%
    \textcopyright{} 2025 IEEE. Personal use of this material is permitted.
    Permission from IEEE must be obtained for all other uses, in any
    current or future media, including reprinting/republishing this
    material for advertising or promotional purposes, creating new
    collective works, for resale or redistribution to servers or lists,
    or reuse of any copyrighted component of this work in other works.
    Published version: \url{https://doi.org/10.1109/BigData66926.2025.11402029}.
}
\endgroup

\begin{abstract}
Decision trees are among the most well-known interpretable models in Machine Learning. 
Over the years, a wide variety of strategies have been proposed to define split conditions within the greedy learning framework, aiming to guide instance routing and enhance predictive performance.
These strategies range from axis-parallel and oblique splits based on feature values to proximity-based splits using exemplar instances and more complex hybrid approaches. 
However, no single method has emerged as universally superior.
In this work, we present two key contributions to the construction of greedy decision trees. 
First, we propose a hybrid tree architecture that integrates both feature-based and case-based splits, allowing for heterogeneous splits throughout the tree. 
Second, we introduce a novel type of split that leverages a selective case-based approach: when a split is needed, our method searches for the most informative subset of features to define the split, effectively balancing the specificity of case-based methods with the simplicity of feature-based ones.
Experiments on a diverse set of tabular datasets show that our method outperforms standard decision trees and achieves performance on par with ensemble methods, while producing shallow trees that retain or even improve predictive accuracy.
\end{abstract}


\section{Introduction}
Machine Learning (ML) models are increasingly used to support critical decision-making tasks, including credit risk assessment, job candidate selection, and medical treatment planning~\cite{bhatore2020machine,DeFauw2018ex1,garg2022review}. 
Despite their strong predictive performance, many of these models rely on complex architectures that hinder their interpretability. 
Addressing this limitation is the goal of Explainable AI (XAI)~\cite{bodria2023benchmarking}.
Among ML models, decision trees (\textsc{dt})~\cite{breiman1984classification} stand out as the most prominent interpretable decision-making tools. 
Their hierarchical rule-based structure offers inherent transparency, facilitating model comprehension by both domain experts and end-users~\cite{Rudin19stop}. 
Moreover, when combined with ensemble techniques, \textsc{dt}s have been shown to outperform deep learning models on tabular data~\cite{Grinsztajn22outperform}.

Different learning strategies for growing \textsc{dt}s have been proposed, the most common being top-down greedy partitioning strategies, as used by well-known algorithms such as CART, ID3, and C4.5~\cite{Rokach2005topdownlearnsurvey}. 
These methods construct \textsc{dt}s in a greedy, recursive manner, producing sub-optimal trees that are computationally efficient to learn. 
Despite their lack of optimality, they offer a practical trade-off by significantly reducing the computational cost compared to the construction of truly optimal decision trees, a task known to have prohibitive computational requirements~\cite{bertsimas2017optimal,Demirovic2022otpimaldynamic}.
Although greedy approaches are the most commonly used and studied in the literature, alternative methods for efficiently constructing sub-optimal \textsc{dt}s have also been explored. 
Some examples include evolutionary methods~\cite{Kretowski2004evol} and generative approaches~\cite{guidotti2024gentree}.

A characteristic that distinguishes different implementations of \textsc{dt}s consists in how they implement splitting criteria of internal nodes. 
Standard implementations employ splits that are axis-parallel to the feature space, leading to \textit{univariate} decision trees (\textsc{udt})~\cite{quinlan1993c45}. 
Other variants introduce oblique splits, which use linear combinations of feature values to define the decision boundaries, resulting in multivariate or \textit{oblique} decision trees (\textsc{odt})~\cite{BrodleyU95multivar}, theoretically more expressive than traditional \textsc{udt}s.
Other methods focuses on case-based \textsc{dt}s that operate within a distance-based feature space rather than the original input feature space~\cite{Chouakria2012tscasebased,Lucas2019proximityfor,yamada2003split,cascione2024pivotree}. 
In such \textsc{dt}s, splitting functions are defined through proximity-based comparisons, where the routing of an instance depends on its distance to one or more exemplar instances. 
The decision criterion may be defined in absolute terms, by checking whether the distance exceeds a predefined threshold, or in relative terms, by comparing the distances of an instance to a pair of exemplars and routing it based on which exemplar it is closer to.

All the approaches discussed above build what we refer to as \emph{homogeneous} \textsc{dt}s, where the same type of split is used throughout the entire tree. 
As a result, all decision paths follow a uniform logic and rely on a fixed \emph{a priori} representation for modeling decision boundaries. 
This structural bias may limit \textsc{dt}s expressiveness, and hence their performance.
To overcome this limitation, hybrid \textsc{dt} approaches allowing \emph{heterogeneous} trees, employing different types of splitting functions within the structure, have been proposed, with examples in both the greedy~\cite{mivskovic2014hybr} and evolutionary~\cite{Llora2004mixedDTinLCS,Kretowski2006evolmixedtree} learning frameworks.
Following this line of research, we propose \textsc{HybridDecisionTree}, a \textsc{dt} structure that leverages heterogeneous feature-based and case-based splitting strategies within a unified, greedy learning framework. 
This approach retains the flexibility typical of hybrid \textsc{dt} models, while maintaining interpretability and consistency by relying solely on a unified rule-based framework.
Furthermore, we introduce a novel form of learnable \textsc{dt} that performs case-based splits guided by the relevance that small groups of features have for case-based comparison. 
We propose \textsc{SelectivePivotTree}, a \textsc{dt} in which each internal node adaptively selects a subset of the most relevant features, which are then used to compute case-based splits, both in absolute and relative terms. 
This relevance-driven selection enables more expressive and targeted decision logic, while preserving interpretability and guiding instance routing through the tree.
Experiments on a wide collection of tabular datasets demonstrate that our proposals perform competitively with state-of-the-art \textsc{dt} methods. 
Also, incorporating \textsc{SelectivePivotTree} into \textsc{HybridDecisionTree} often results in better performance than both homogeneous \textsc{dt}s and ensembles of homogeneous \textsc{dt}s.

The rest of the paper is structured as follows. 
In Section~\ref{sec:related}, we review related work, while in Section~\ref{sec:background}, we introduce the necessary background and formalize the \textsc{dt} framework used throughout the paper.
In Section~\ref{sec:method}, we introduce our proposals, and in Section~\ref{sec:experiments}, we report the experimental results. Finally, Section~\ref{sec:conclusion} summarizes the contributions and details future research directions.

\section{Related Works}
\label{sec:related}
In the following, we propose a categorization of \textsc{dt}s learning approaches that integrate different splitting strategies within a single tree, and examine related methods that use subset-based feature selection in case-based splits.

Hybrid approaches to model learning have been extensively explored in the ML literature as effective strategies for overcoming the limitations of individual methods, while enhancing both predictive performance and computational efficiency~\cite{Azevedo2024hybridreview,wozniak2013hybrid}.
In~\cite{mivskovic2014hybr}, the authors focus on hybrid \textsc{dt}-based methods. 
We can broadly distinguish between \emph{leaf}-oriented and \emph{structure}-oriented hybrid \textsc{dt} approaches. 
\emph{Leaf}-oriented hybrid approaches focus on partitioning the feature space using standard splitting strategies, and then replacing the leaves with models based on different classifiers. 
For instance,~\cite{Utgoff88perceptrontree} uses linear threshold units,~\cite{frank1998using} linear regressions, while models described in~\cite{liu20017logistic} employ logistic regression.
Similarly,~\cite{siang2011naivebayesrtree} and~\cite{wang2024generalnaive} employ Naive Bayes and General Naive Bayes classifiers, while~\cite{islam2023knntree} uses $k$-Nearest Neighbors. 
In~\cite{Kumar2010svmdt}, Support Vector Machines (\textsc{svm}s) are used as leaf classifiers within a \textsc{dt} trained to distinguish instances based on their proximity to the \textsc{svm} decision boundary, while in~\cite{ZhouC02HDT} neural networks are used for non ordinal attributes at leaves.
\emph{Structure}-oriented hybrid approaches allow for non-standard split conditions throughout the entire tree structure. 
Functional trees~\cite{gama2004functional} introduce nodes where original input features are evaluated along with newly constructed attributes derived from constructor functions trained locally at each node. 
The authors use generalized linear models~\cite{nelder1972generalized} as constructors, resulting in new attributes corresponding to linear combinations of the original input features, defining a multivariate decision surface.
On the other hand, in~\cite{Guo1992featext} the authors employ a multilayer perceptron at each node to extract nonlinear features to be used for splitting, while in~\cite{carson2020hybrid} neural networks' outputs are used at each node to route instances in the tree. 
Evolutionary approaches for learning heterogeneous \textsc{dt}s have also been proposed, focusing on univariate and multivariate feature-based splits~\cite{Llora2004mixedDTinLCS,Kretowski2006evolmixedtree}.

Our proposal belongs to \emph{structure}-oriented hybrid \textsc{dt} strategies, and extends prior works in multiple directions. 
Unlike many of the previous methods that rely on external classifiers to define non-standard split conditions, our approach evaluates both feature-based and case-based rules without requiring auxiliary classifiers, allowing heterogeneous split conditions to be selected independently at each internal node. 
The result is a more interpretable model structure that offers greater flexibility while remaining easier to use and understand. 
Furthermore, incorporating case-based split conditions to the tree structure broadens the types of decision rules that can be applied within the tree, providing greater flexibility for different tasks. 

In our proposal, we introduce a splitting strategy inspired by shapelets~\cite{ye2009shapelets}, which focuses on identifying relevant subsets of features for case-based splits. 
Thus, we briefly review notable related approaches in the following.
In the field of time-series, \textsc{dt}s employing shapelet-based split criteria have been studied since the introduction of shapelets. 
In~\cite{Ye2011shapelttree}, the authors propose \textsc{dt}s in which instances are routed based on their similarity to a selected shapelet at each node, where a shapelet refers to a discriminative subsequence of a time-series that is informative for classification tasks.
Shapelets are identified in~\cite{Ye2011shapelttree} through a brute-force strategy.
To improve computational efficiency, of such approaches, randomized ensemble methods have been proposed as extensions~\cite{Karlsson2015forest,Karlsson2016gRSF}.
In~\cite{Yuan2022pairwiserandomshaplet}, a pairwise ensemble approach is introduced, where routing decisions on single trees are made based on comparative proximity to shapelets associated with each branch. 
Also, in~\cite{Hills2014shaptransf}, the authors propose transforming time-series data into a new shapelet-based feature space where each feature encodes the proximity of an instance to a specific shapelet. 
Classifiers, including interpretable ones such as \textsc{udt}s or \textsc{odt}s, can then be trained on this transformed representation.
A notable example of a hybrid decision tree is \textsc{JS48SS}~\cite{Brunello2019js48ss}, which combines shapelet-based and feature-based splits to handle both tabular and sequential data, by leveraging discriminative closed frequent patterns for sequence analysis and evolutionary algorithms for time-series classification, resulting in a heterogeneous tree structure in terms of both decision logic and data-type-specific processing.

In contrast to prior approaches, we exploit the shapelet-based paradigm into the tabular domain by selecting informative subsets of features to define case-based splits. 
These splits can be integrated into the hybrid \textsc{dt} framework, unifying both feature-based and case-based decision logic.

\section{Setting the Stage}
\label{sec:background}
To keep our paper self-contained, we report here a brief overview of concepts necessary to comprehend our proposal.
\textsc{dt}s can be considered a special case of rule-based (\textsc{rb}) models.
\textsc{rb} models are defined in terms of a set $\Phi = \{\phi_1, \dots, \phi_r\}$ of \emph{rules}, where each rule defines a set of constraints and a label: whenever the constraints are verified, then the label holds.
Formally, a rule $\phi = (g^\phi, y^\phi)$ is a pair comprised of a list of indicator, constraint, or splitting functions $g^\phi = \{g^\phi_1, \dots, g^\phi_q \}$ and a label $y^\phi$. 
Thus, we have that $g_i^\phi(x) = 1$ when the condition expressed by $g_i^\phi$ is respected by $x$, $g_i^\phi(x) = 0$ otherwise.
Using $\mathbbm{1}(\mathit{cond})$ to refer to a generic indicator function with a condition $\mathit{cond}$ we can write in general $g_i^\phi = \mathbbm{1}(\phi_i(x))$ to refer to the verification of the $i$-th condition of $\phi$ on $x$.
A record $x$ respects a rule $g^\phi$, i.e., $g^\phi = \mathbbm{1}(\phi(x)) = 1$, when all the constraints $g_i^\phi$ are respected by $x$, i.e., $\sum_{i=1}^q \mathbbm{1}(\phi_i(x)) = q$.

\textsc{dt}s models share much of their structure with \textsc{rb} models. 
Indeed, also \textsc{dt}s induce a set of indicator functions, which are laid in a tree-like hierarchy.
Elements in the tree are called \emph{nodes}, with the starting node being the \emph{root} of the tree, and terminating nodes, in which rule labels are stored, being the \emph{leaves} of the tree.
Root-to-leaf paths define conjunctions of indicator functions, representing rules.
Paths of equal length define \textit{levels}, and the maximum level defines the \textit{depth} of the tree.
Thus, inference is the process of identifying the unique root-to-leaf path, i.e., the unique rule $\phi$, that is verified by an instance $x$.
Given a \textsc{dt} $f$, inference on $x$ can defined as
\begin{equation*}
    f(x) = \sum_{\phi \in \Phi} \mathbbm{1}(\phi(x))y^{\phi}
\end{equation*}

For each split defined by $g_i^\phi(x)$ of $f$, let $X_i$ be a set of instances during the training of $f$ at iteration $i$.
We can define the \textit{left} and \textit{right} subsets of the split as:
\begin{equation}
\label{eq:subsets}
    X_l {=} \{ x \in X_i \mid g_i^\phi(x) = 1 \}, \;
    X_r {=} \{ x \in X_i \mid g_i^\phi(x) = 0\}
\end{equation}
and the \textit{information gain} $\Delta_i$ of the indicator $g_i^\phi(x)$ as:
\begin{equation}
\label{eq:information_gain}
    \Delta_i = \mathcal{I}_{g_i}(X_i, Y_i) - \frac{|X_l|}{|X_i|} \mathcal{I}(X_l, Y_l) - \frac{|X_r|}{|X_i|} \mathcal{I}(X_r, Y_r)
\end{equation}
where $\mathcal{I}$ is a measure of impurity or uncertainty w.r.t class labels, such as Shannon entropy or the Gini index~\cite{tan2006data}.
Thus, the selection of the best indicator function
$g_i^\phi$ depends on the information gain $\Delta_i$, that in turns depends on how the indicator function $g_i^\phi$ is implemented and by its parameters.

\smallskip
A key distinguishing feature among different \textsc{dt} models lies in the implementation of indicator functions $g_i^\phi$.
This, in turn, influences the expressive power of the tree and shapes its underlying decision logic. 
Given an indicator function $g_i^\phi$ associated to a non-leaf node, we can broadly distinguish between \emph{feature-based} and \emph{case-based} indicator functions, in the rest called ``splitting functions'' for ease of exposition.

\emph{Feature-based} splitting functions represent the basic implementation of \textsc{dt}s. These trees employ functions of the form:
\begin{equation}
\label{eq:feat_split}
    g_i^\phi(x) = \mathbbm{1} \left( \alpha_i^\intercal x \leq \beta_i\right)
\end{equation}

\noindent where $x\in \mathbb{R}^m$ is the instance to route, described by a total of $m$ features, and $\alpha_i \in \mathbb{R}^m$ and $\beta_i \in \mathbb{R}$ are parameters of $g_i^\phi$.
In line with standard practice, we always consider real-valued input data in our discussion. 
Most greedy, top-down \textsc{dt} learning strategies estimate $\alpha_i$ and $\beta_i$ by selecting the values that best partition the instances routed at $g_i^\phi$ associated non-leaf node from previous splits, according to a chosen impurity or information-based criterion~\cite{breiman1984classification}.
Therefore, different splitting strategies aim to find parameters $\alpha_i, \beta_i$ that maximize the information gain $\Delta_i$ over all admissible choices of $\alpha$ and $\beta$, i.e., $\alpha_i, \beta_i = \arg\max_{\alpha, \beta} \, \Delta_i(\alpha, \beta)$.

Feature-based splits differ in terms of which $\alpha_i$ they consider admissible.
\textit{Univariate} decision trees (\textsc{udt}s) implement axis-parallel splits, meaning that each component of $\alpha_i$ is zero except for a single component which is $1$, i.e., $\sum_{j=1}^m \alpha_{i,j} = 1$.
Different strategies for learning \textsc{udt}s have been proposed~\cite{Kotsiantis2013overview,tan2006data}.
In this work, we assume the standard CART strategy~\cite{breiman1984classification,tan2006data} when applying univariate splits.
In contrast, \textit{oblique} decision trees (\textsc{odt}s) allow for multiple non-zero components in $\alpha_i$, and for arbitrary value of such components, resulting in linear combinations of feature values. 
This enables splits oblique to the axis. 
Different strategies for learning the best feature combinations for splitting have been proposed~\cite{BrodleyU95multivar,canete2021review}.
In this work, we adopt the HHCART strategy~\cite{Wickramarachchi2016HHCART}, restricting $\alpha_i$ to have exactly two non-zero components to avoid complex rules that could hinder interpretability.

\emph{Case-based} splits, in contrast, shift the focus of the splitting function to instance comparison w.r.t. exemplar training instances, referred to as \emph{pivots}~\cite{cascione2024pivotree,Lucas2019proximityfor,Yuan2022pairwiserandomshaplet}.
We further categorize these splits into \emph{distance} and \emph{proximity}-based splits.

\emph{Distance}-based splits use splitting functions of the form:
\begin{equation}
\label{eq:dist_split}
    g_i^\phi(x) = \mathbbm{1} \left( d(\alpha_i^\intercal x, \alpha_i^\intercal p) \leq \beta_i \right)
\end{equation}
where $d : \mathbb{R}^m \times \mathbb{R}^m \rightarrow \mathbb{R}$ is a distance function quantifying the resemblance between instances, and $p$ is a reference pivot instance.
This enables decision rules of the form: ``if the distance between instance $x$ and pivot $p$ is greater than or equal to threshold $\beta_i$, then...''. 
Pivots are typically selected from the training data either randomly or as those instances that best discriminate between classes. 
Examples of such approaches include \textsc{PivotTree} (\textsc{pt})~\cite{cascione2024pivotree}, which identifies, for each internal node, a pair of descriptive and discriminative instances and selects the best combination of $p$ and $\beta_i$ according to an information gain criterion. 
Thus, since such approaches operate directly in the original feature space, 
each $\alpha_i$ is always fixed to the all-ones vector, i.e., $\sum_{j=1}^m \alpha_{i,j} = m$. 
Additionally, the authors in~\cite{Fedele2025Interpretable} address the problem of interpretability in distance-based models applied to high-dimensional datasets, demonstrating that learning $d$ as \textsc{udt} enables easy-to-interpret distance-based models with virtually no loss in performance.

\emph{Proximity}-based splits, 
is a more general form of a distance-based split that employs splitting functions of the form:
\begin{equation}
\label{eq:prox_split}
    g_i^\phi(x) = \mathbbm{1} \left( d(\alpha_i^\intercal x, \alpha_i^\intercal p_{-}) \leq d(\alpha_i^\intercal x, \alpha_i^\intercal p_{+}) \right)
\end{equation}
where $p_{-}$ and $p_{+}$ are two reference pivots. 
This type of split compares the proximity of the instance $x$ to both pivots, enabling reasoning of the form: ``if instance $x$ is closer to pivot $p_{-}$ than to pivot $p_{+}$, then...''. 
This approach relies on relative distance comparisons rather than absolute thresholds. 
Examples of \textsc{dt}s adopting such strategy are found in~\cite{Chouakria2012tscasebased}, and, in an ensemble setting, in~\cite{Lucas2019proximityfor} and~\cite{Yuan2022pairwiserandomshaplet}, all introducing case-based trees with a focus on time-series classification. 
Also in this case, for each $\alpha_i$ we adopt an all-one vector.

We emphasize that, consistently with feature-based splitting functions, the greedy tree construction using distance-based or proximity-based procedure remains unchanged.  
Indeed, also in the case of case-based splitting functions, at each iteration $i$, the greedy algorithm selects $\alpha_i \in \mathbb{R}^m$ and $\beta_i \in \mathbb{R}$, along with the pivot instances, that maximize the information gain $ \Delta_i$.

\begin{figure*}[t]
    \centering
    \includegraphics[width=0.20\linewidth]{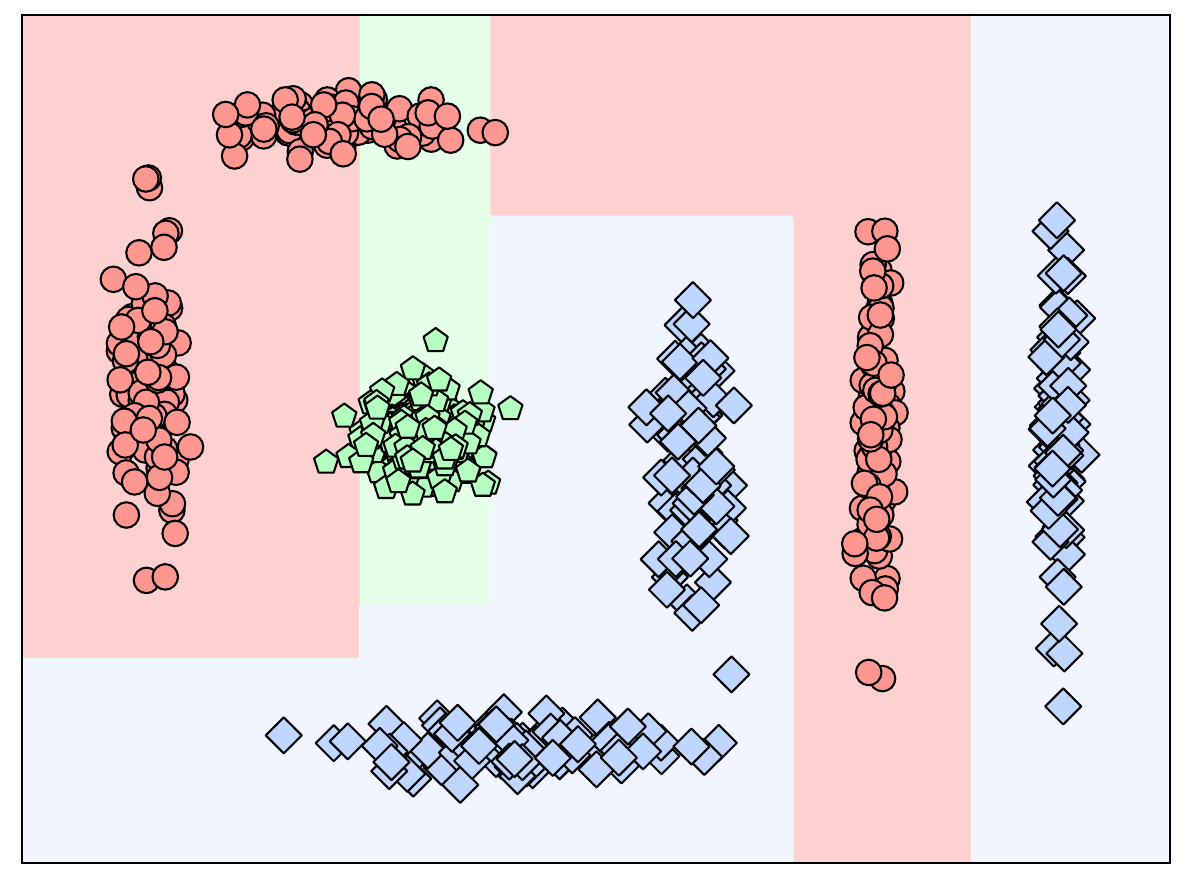}
    \includegraphics[width=0.20\linewidth]{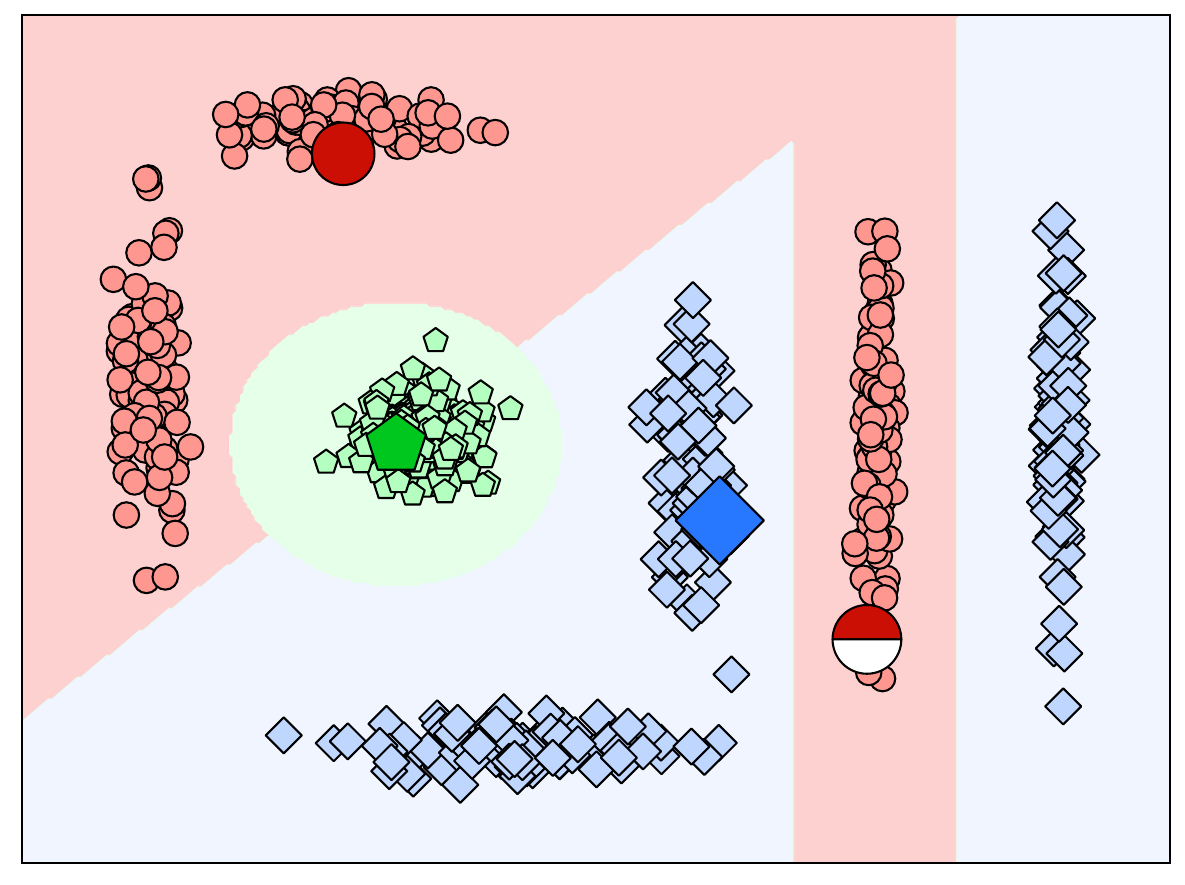}
    \includegraphics[width=0.31\linewidth]{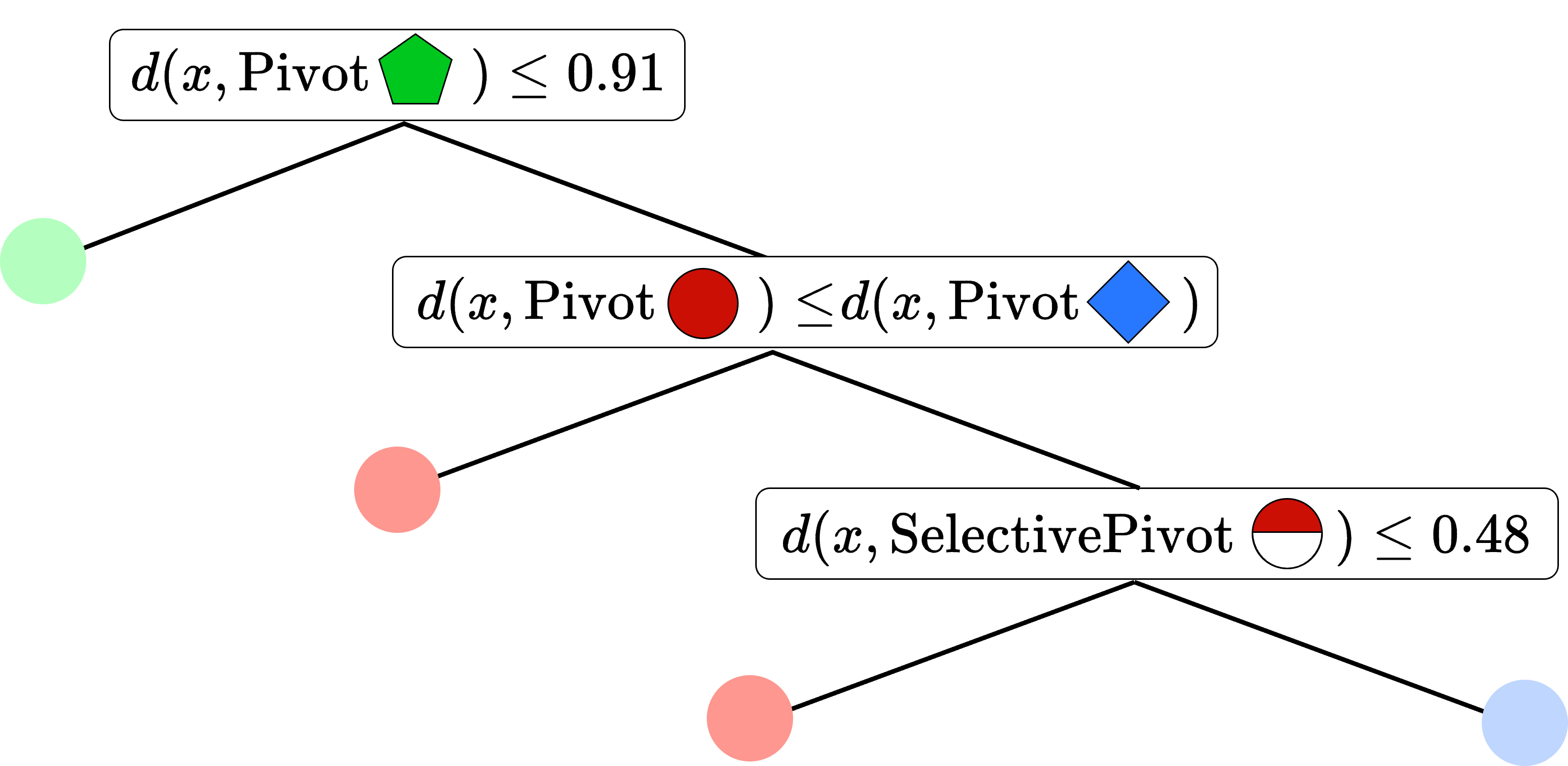}
    \caption{Decision boundaries of a univariate \textsc{dt} (left), and of a hybrid \textsc{dt} (center), both trained on a synthetic dataset with $\mathit{max\_depth}=3$. 
    The hybrid \textsc{dt} (right) shows a distance-based split, a proximity-based split, and a selective distance-based split. 
    Pivots are highlighted with bigger size in the central figure.}
    \label{fig:hybrdi_tree}
\end{figure*}

\section{Methodology}
\label{sec:method}
In this section, we first introduce \textsc{HybridDecisionTree}, a greedy learning framework for constructing \emph{heterogeneous} \textsc{dt}s that allow different types of splitting functions within the same tree. We then present \textsc{SelectivePivotTree}, a case-based \textsc{dt} that employs a splitting strategy based on the selection of informative feature subsets for case-based routing. 
We underline that, in this paper, we focus on \textsc{dt}s trained using a greedy strategy, where candidate splits are evaluated using Gini index as impurity measure $\mathcal{I}$, following the classical CART strategy~\cite{breiman1984classification}. 
We complement the section with a theoretical analysis demonstrating the advantages of our proposal over traditional \textit{homogeneous} \textsc{dt}s.

\smallskip
\textbf{Hybrid Decision Tree.}
We present here \textsc{HybridDecisionTree} (\textsc{hdt}) , an interpretable tree-based decision model leveraging heterogeneous splits in its structure.
\textsc{hdt} adopts a greedy, top-down induction strategy analogous to standard \textsc{dt} learning.
Given the premises in Section~\ref{sec:background}, constructing a \textsc{dt} that integrates the different splitting strategies into an hybrid way seems quite straightforward.
However, while various existing \textsc{dt} approaches allowing hybrid splits have been proposed in the literature combining heterogeneous types of splits within the same tree structure~\cite{Brunello2019js48ss,Kretowski2006evolmixedtree,mivskovic2014hybr}, none of them integrates both feature-based and case-based splits for tabular data within a greedily learned framework. 

The key difference between a \textsc{hdt} and a standard \textsc{dt} is that, besides the labeled training data $\langle X, Y \rangle$, the algorithm also receives as input a set $\Gamma$ of types of heterogeneous splitting function strategies.
In this paper, we aim to demonstrate that combining heterogeneous decision logics in $\Gamma$, such as feature-based and case-based splits, can be beneficial across different contexts, improving performance without sacrificing interpretability, while often yielding shallower trees.

More in detail, the learning process of \textsc{hdt} is described by
Algorithm~\ref{alg:hybrid-tree} and works as follows.
Algorithm~\ref{alg:hybrid-tree} revolves around a recursive \textsc{hdt} procedure named \textsc{hdtr} (line 1).
The \textsc{hdtr} procedure is described in lines 3--16: if the current depth of the tree \textsc{depth}$(T)$ is lower than the maximum tree depth $\mathit{max\_depth}$ (line 4), then the set of candidate indicator functions $G$ and the set of their respective information gains $D$ are initialized as empty sets (line 5). 
Then, for each candidate splitting strategy $\gamma_i \in \Gamma$, the \textsc{MakeSplit} function identifies the optimal indicator function $g_i$ that maximizes the information gain $\Delta_i$, as defined in Eq.~\ref{eq:information_gain}, based on the specific logic and mechanics of the splitting strategy $\gamma_i$ (lines 6--8). 
For instance, if $\gamma_i$ corresponds to a feature-based univariate splitting strategy, \textsc{MakeSplit} identifies the best univariate indicator function $g_i$ using a standard criterion such as CART, consistent with Eq.~\ref{eq:feat_split}. 
This implies selecting the optimal parameters $\alpha_i$ and $\beta_i$ to perform an axis-parallel split.
On the other hand, if $\gamma_i$ encodes a distance-based, case-based splitting strategy, \textsc{MakeSplit} selects the best indicator function $g_i$ by evaluating distance comparisons between instances and candidate pivots, in accordance with Eq.~\ref{eq:dist_split}, determining the best combination of $\alpha_i$, $\beta_i$, and pivot $p$ according to the chosen case-based splitting strategy.
An analogous procedure is followed for all other splitting strategies described in Section~\ref{sec:background}.
Then, each indicator function $g_i$ and information gain $\Delta_i$ are added to results sets $G$ and $D$, respectively (line 8). 
Among the indicator functions in $G$ obtained through different heterogeneous splitting strategies, \textsc{hdtr} selects the one having the highest information gain to be used for the split (line 9).

\begin{algorithm}[t]
\footnotesize
\caption{\textsc{HybridDecisionTree}$(X, Y, \Gamma)$}
\label{alg:hybrid-tree}
\begin{algorithmic}[1]
    \Input $\langle X, Y \rangle$ dataset, $\mathit{max\_depth}$ max depth, $\Gamma$ splitting function types.
    \Output $T$ learned tree
    \State $T \gets \Call{hdtr}{\emptyset, X, Y, \Gamma}$;
    \State \Return $T$
    \newline
    
    \Function{hdtr}{$T, X, Y, \Gamma$}
    \If{$\Call{depth}{T} \leq \mathit{max\_depth}$}
        \State $G \gets \emptyset$; $D \gets \emptyset$;
        \For{$\gamma \in \Gamma$}
            \State $g_i, \Delta_i \gets \Call{MakeSplit}{X, Y, \gamma}$;
            \State $G \gets G \cup \{g_i\}$; $D \gets D \cup \{\Delta_i\}$;
        \EndFor
        \State $g^* \gets \arg\max_{g_i \in G} \Delta_i$;
        \State $X_l, X_r, Y_l, Y_r \gets \Call{SplitData}{X, Y, g^*}$;
            
        \State $T_l \gets \Call{hdtr}{T, X_l, Y_l, \Gamma}$;
        \State $T_r \gets \Call{hdtr}{T, X_r, Y_r, \Gamma}$;
        \State $T \gets \Call{AddSplitToTree}{T, T_l, T_r}$;
        \State \Return $T$
        \Else
            \State \Return $\Call{makeLeaf}{T}$
        \EndIf
    \EndFunction
    
\end{algorithmic}
\end{algorithm}

The best indicator function $g^*$ is then used to partition the data with the \textsc{SplitData} function, obtaining left and right sets $X_l, X_r$ as described in Eq.~\ref{eq:subsets}, with their respective labels $Y_l,Y_r$ (line 10).
Subsequently, \textsc{hdtr} recursively processes the left and right subsets, $\langle X_l, Y_l \rangle$ and $\langle X_r, Y_r \rangle$, respectively, and composes the resulting subtrees (lines 11--14). 
Conversely, if the maximum depth is reached (line 15) or other stopping criteria here not specified are satisfied, the recursion terminates and a leaf node is returned (line 16).

Figure~\ref{fig:hybrdi_tree} shows the decision boundaries induced by a univariate \textsc{dt} (left) and by a \textsc{hdt} (center), on a synthetic dataset with two features and three classes. 
Both models are trained with a maximum depth of $\mathit{max\_depth} = 3$.
We also report a visual depiction of the trained \textsc{hdt} for completeness.
The decision logic of the trained \textsc{hdt} can be interpreted as follows: a test instance $x$ is first compared to the green pivot using a distance-based split. 
If the distance is less than or equal to $0.91$, the instance is assigned to the green class. 
This split is depicted as a green circular region of radius $0.91$ centered on the selected pivot instance, shown as a large green pentagon.
If $x$ is too far from the green pivot, a proximity-based split is performed: if the instance is closer to the red circle than to the blue diamond, it is assigned to the red class. 
If not, a further comparison is made using the new distance-based selective strategy described in the next section. 
We anticipate to the reader that such strategy compares an instance $x$ to a pivot, depicted as a half-red, half-white circle in Figure~\ref{fig:hybrdi_tree}, by computing the distance to such pivot using only the first feature, while disregarding the second one.
We underline how such an heterogeneous structure allows \textsc{hdt} to form fine-grained decision boundaries while maintaining interpretability through a compact tree composed of only a few splits.
In contrast, a standard univariate \textsc{dt} of the same depth is unable to fully separate the data points belonging to different classes.

As anticipated by the previous example, besides the standard feature-based and case-based splits, \textsc{hdt} can also incorporate a variant of case-based splits, where the distance $d$ between an instance $x$ and a pivot $p$ is computed using only a relevant subset of features.
Details are provided in the following.

\begin{algorithm}[t]
\footnotesize
\caption{\textsc{MakeSelectiveSplit}$(X, Y, d, v, h, k, \pi)$}
\label{alg:selective-split}
\begin{algorithmic}[1]
    \Input $\langle X, Y \rangle$ dataset,  
    $d$ distance function, 
    $v$ fraction of instances,
    $h$ nbr. of features to select, 
    $k$ nbr. of candidate pivots,
    $\pi$ use proximity split flag.
    \Output $g^*,\Delta^*$ best splitting function and corresponding information gain.
    \State $X' \gets \Call{CandidateInstances}{X,v}$;
    
    \State $A \gets \Call{SelectableFeatures}{X', h}$;
    \State $P \gets \Call{CandidatePivots}{X', Y, A, k}$;
    \State $g^*\gets \emptyset$, $\Delta^*\gets -\infty$;
    
 \For{$\alpha_i \in A$}
    \If{${\pi}$}
        \State $g_{i} \gets \Call{PivotSplit}{\alpha_i^\intercal X, Y, \alpha_i^\intercal P, d}$ using Eq.~\ref{eq:dist_split};
    \Else
        \State $g_{i} \gets  \Call{ProximitySplit}{\alpha_i^\intercal X, Y, \alpha_i^\intercal P, d}$ using Eq.~\ref{eq:prox_split};
    \EndIf
    \State $\Delta_i \gets \mathcal{I}_{g_i}(\alpha_i^\intercal X, Y)$;
    \If{$\Delta_i > \Delta^*$}
        \State $g^* \gets g_i$, $\Delta^* \gets \Delta_i$;
    \EndIf
\EndFor

\State \Return $g^*,\Delta^*$

\end{algorithmic}
\end{algorithm}

\smallskip
\textbf{Selective Pivot Tree.}
In this section we introduce \textsc{SelectivePivotTree} (\textsc{spt}), a \textsc{dt} variant that performs case-based splits based on distances computed over \textit{selected} subsets of features. 
A \textsc{spt} is learned greedily, similar to standard top-down case-based learning approaches~\cite{cascione2024pivotree}. 
The key difference lies in the splitting logic at the internal nodes.
Inspired by the shapelet paradigm~\cite{ye2009shapelets}, \textsc{spt} employs splits that compare an instance $x$ to a pivot $p$ only with respect to a relevant subset of features $h < m$.
Our proposal extends traditional case-based splits~\cite{cascione2024pivotree} by constraining the learnable vector $\alpha_i$ used in Eq.~\ref{eq:dist_split} and Eq.~\ref{eq:prox_split} to be a binary vector having exactly $h \geq 1$ components of value $1$, with all others set to zero, i.e., $\sum_{j=1}^m \alpha_{i,j} = h < m$ with $\alpha_i \in \{0,1\}^{m}$.
Once $\alpha_i$ is defined, the distance of $x$ to pivot $p$ is computed as $d(\alpha_i^\intercal x, \alpha_i^\intercal p)$, such that only the components used are those equals to one in $\alpha_i$, thus impacting the distance \mbox{between $x$ and $p$.} 

Algorithm~\ref{alg:selective-split} outlines the \textsc{MakeSelectiveSplit} strategy.
Given data and labels $\langle X, Y \rangle$, the distance function $d$, the fraction of initial instances $v$, the number of features to select $h$, the number of candidate pivots to consider $k$, and $\pi$ a 
flag indicating the type of case-based split to use, i.e., proximity-based if $\pi = \mathit{True}$, distance-based otherwise.
Like any \textsc{MakeSplit} function, \textsc{MakeSelectiveSplit} returns the best indicator $g^*$ and its corresponding gain $\Delta^*$.

Algorithm~\ref{alg:selective-split} starts by selecting a subset of training instances $X' \subseteq X$ via random sampling of $\lceil v |X| \rceil$ elements from $X$, where $v \in (0,1]$ with \textsc{CandidateInstances} (line 1).
These instances are retained for the following steps.
If $v = 1$, all instances are considered.
Next, \textsc{SelectableFeatures} generates a set $A$ of binary vectors $\alpha_i \in \{0,1\}^m$ with exactly $h$ nonzero entries, i.e., $\sum_{j=1}^m \alpha_{i,j} = h < m$ (line 2).
Instead of exhaustively exploring all $\sum_{i=1}^m \binom{m}{i}$ combinations, we implement and compare four versions of \textsc{SelectableFeatures}:
\begin{itemize}[noitemsep,topsep=0pt]
    \item \textit{Fixed-Combinations (\textsc{fc}):} Considers the selection of $\binom{m}{h}$ combinations of $h$ features among $m$ alternatives.
    \item \textit{Feature-Importance:} Importance scores are computed from $\langle X', Y \rangle$ using simple models (e.g., linear or Random Forest), ranked, and the top $h < m$ features are selected.
    All $\sum_{i=1}^h \binom{h}{i}$ combinations over these are then considered.
    The value $h$ is determined via:
    \begin{itemize}[noitemsep,topsep=0pt]
         \item \textit{Elbow (\textsc{fie}):} Applies a clustering-inspired elbow~\cite{tan2006data} method assuming monotonic decrease in importance.
         \item \textit{Drop (\textsc{fid}):} Picks $h$ at the point of the steepest drop in consecutive importance scores~\cite{guidotti2015tosca}.
    \end{itemize}
    \item \textit{Greedy-Combinations (\textsc{gc}):} Importance scores are repeatedly computed over $\langle X', Y \rangle$, excluding previously selected top features.
    At each iteration $j$, considers the combination of the most important $j$ features, and the process continues until $j \leq \lfloor \frac{-1 + \sqrt{1+8m}}{2} \rfloor$.
\end{itemize}
    
Then, the set $P \subseteq X'$ of $k$ candidate pivots is extracted using the \textsc{CandidatePivots} function (line 3) for which we design and test three different variants: 
\begin{itemize}[noitemsep,topsep=0pt]
    \item \textit{All (\textsc{a}):} Selects all instances as candidate pivots, i.e., $P=X'$. The hyper-parameter $k$ is not used.
    
    \item \textit{Random (\textsc{r}):} Selects candidate pivots by randomly sampling $k$ instances from the training set $X'$.
    
    \item \textit{Clustering (\textsc{c}):} $k$-Medoids clustering algorithm~\cite{park2009kmedoids} is executed on $\alpha_i^\intercal X'$ for each $\alpha_i \in A$ to identify $k$ clusters that are homogeneous w.r.t. the selected components in $A$. 
    The resulting $k$ medoids, i.e., the instances that minimize the total dissimilarity within each cluster, are selected. 
    
\end{itemize}

\smallskip
After this, the best indicator function $g^*$ and best information gain $\Delta^*$ are initialized as an empty set and $-\infty$, respectively (line 4).
Consequently, the search for the optimal indicator function is started, as described in lines 4--11. 
For each selected features $\alpha_i \in A$ (line 5), if the flag $\pi$ is set to \textit{True} (lines 6--7), a distance-based strategy is employed; otherwise (line 8--9), a proximity-based strategy is used to determine the most informative split associated with $\alpha_i$.

More in detail, when $\pi = \mathit{False}$, \textsc{MakeSelectiveSplit} computes the distance-based split by \textsc{PivotSplit} as in~\cite{cascione2024pivotree}, (line 7) thus implementing a \textsc{SelectivePivotTree} (\textsc{spt}).
Here, we consider the result of computing the distance $d$ of each instance $x \in X $ w.r.t. each one of the candidates $p \in P$, collecting distance values for each instance in a real-valued vector but only considering the non-zero features in $\alpha_i$, i.e., $d(\alpha_i^\intercal x, \alpha_i^\intercal p)$.
In this way, we represent the instances of $\alpha_i^\intercal X$ in a feature space where each dimension stands for the distance an instance has to a candidate pivot $\alpha_i^\intercal p$ in terms of the features selected by $\alpha_i$. 
From this the name \textit{selective} pivots.
In this data representation, a univariate split can be learned by identifying the optimal feature-threshold pair, i.e., $p, \beta_i$, that maximizes the information gain $\Delta_i$ which corresponds to selecting the candidate pivot that best separates $X$ based on a distance-based criterion, according to Eq~\ref{eq:dist_split}.
Similarly, when $\pi = \mathit{True}$, inspired by~\cite{Lucas2019proximityfor}, \textsc{MakeSelectiveSplit} computes with \textsc{ProximitySplit} a modified strategy from the distance-based one proposed above, and thus implementing a \textsc{ProximitySelectivePivotTree} (\textsc{pspt}).
In particular, from $P$, the pair $p_-, p_+$ leading to the best partition of $X$ in terms of $\Delta_i$ is selected, according to the indicator function definition expressed in Eq.~\ref{eq:prox_split}, still considering the selected features $\alpha_i$ in the distance computation. 
The main difference with \textsc{PivotSplit} is that \textsc{ProximitySplit} compares the parameterized distance of an instance to the respective pivots, rather than splitting based on a single distance value.

Then, \textsc{MakeSelectiveSplit} computes the information gain (line 10) and checks whether the current split yields the highest gain observed so far (line 11).
If so, the indicator function $g_i$ that produces the highest gain is assigned to $g^*$, and the corresponding gain value $\Delta^*$ is updated to $\Delta_i$ (line 12).
Once all selectable feature combinations in $A$ have been evaluated, the best splitting function along with its associated information gain is returned (line 13).

\begin{table*}[t]
    \caption{Datasets info: training ($\mathit{tr}$) and test set ($\mathit{ts}$) sizes, $m$ number of features, $c$ number of classes.}
    \label{tab:datasets_info}
    \setlength{\tabcolsep}{1.12mm}
    \begin{tabular}{c|cccccccccccc|ccccccccccccccc}
    \toprule
    & \multicolumn{12}{|c|}{\texttt{small}} & \multicolumn{13}{c}{\texttt{large}} \\
    
    & \rotatebox{90}{\texttt{iris}} 
    & \rotatebox{90}{\texttt{vert}} 
    & \rotatebox{90}{\texttt{home}} 
    & \rotatebox{90}{\texttt{auct}} 
    & \rotatebox{90}{\texttt{seeds}} 
    & \rotatebox{90}{\texttt{ecoli}} 
    & \rotatebox{90}{\texttt{bkmrk}} 
    & \rotatebox{90}{\texttt{magic}} 
    & \rotatebox{90}{\texttt{diab}} 
    & \rotatebox{90}{\texttt{titan}} 
    & \rotatebox{90}{\texttt{glass}} 
    & \rotatebox{90}{\texttt{cover}} 
    & \rotatebox{90}{\texttt{page}} 
    & \rotatebox{90}{\texttt{wine}} 
    & \rotatebox{90}{\texttt{alg}} 
    & \rotatebox{90}{\texttt{comp}} 
    & \rotatebox{90}{\texttt{vehi}} 
    & \rotatebox{90}{\texttt{eye}} 
    & \rotatebox{90}{\texttt{sylv}} 
    & \rotatebox{90}{\texttt{heloc}} 
    & \rotatebox{90}{\texttt{taiw}} 
    & \rotatebox{90}{\texttt{germ}} 
    & \rotatebox{90}{\texttt{brst}} 
    & \rotatebox{90}{\texttt{ion}} 
    & \rotatebox{90}{\texttt{adult}} \\
    
    \midrule
    $\mathit{tr}$ & 120 & 248 & 393 & 1634 & 168 & 268 & 1600 & 268 & 614 & 712 & 171 & 1600 & 4378 & 5197 & 194 & 5771 & 676 & 1600 & 1600 & 8367 & 24k & 800 & 455 & 280 & 24k \\
    $\mathit{ts}$ & 30  & 62  & 99  & 409  & 42  & 68  & 400  & 68  & 154 & 179 & 43  & 400  & 1095 & 1300 & 49  & 1443 & 170 & 400  & 400  & 2092 & 6000  & 200 & 114 & 71  & 6144 \\
    $\mathit{m}$  & 4   & 6   & 7   & 7    & 7   & 7   & 7    & 7   & 8   & 8   & 9   & 10   & 10   & 11   & 12  & 14  & 18  & 20   & 20   & 23   & 23    & 26  & 30  & 34  & 34  \\
    $\mathit{c}$  & 3   & 3   & 2   & 2    & 3   & 8   & 2    & 8   & 2   & 2   & 6   & 2    & 5    & 13   & 2   & 2   & 4   & 2    & 2    & 2    & 2     & 2   & 2   & 2   & 2  \\
    \bottomrule
    \end{tabular}
\end{table*}

\smallskip
\textbf{Advantages of Hybrid Trees.}
We demonstrate here that a \textsc{hdt}, by selecting among heterogeneous splitting strategies, cannot yield internal nodes with lower information gain than a standard homogeneous \textsc{dt}, provided both use the same greedy strategy and impurity measure. 
\textcolor{black}{We underline that our aim is not to introduce a novel theoretical result, but to formally clarify that heterogeneous splits offer a richer search space under the same induction constraints.} 

To begin, it is important to note that Eq.~\ref{eq:feat_split} remains unchanged when transitioning from a standard univariate feature-based split, where $\sum_{j=1}^m \alpha_{i,j} = 1$, to a selective feature-based split, where $\sum_{j=1}^m \alpha_{i,j} \geq 1$. 
The distinction lies in the number of non-zero elements in $\alpha_i$.
Then, a feature-based splitting function with all $\alpha_{i,j} = 1$, i.e., $\sum_{j=1}^m \alpha_{i,j} = m$, is a special case of the distance-based split in Eq.~\ref{eq:dist_split}, where the pivot $p$ is an instance for which $d(\alpha_i^\intercal x, \alpha_i^\intercal p) = \alpha_i^\intercal x$.
Continuing, the distance-based split in Eq.~\ref{eq:dist_split} corresponds to the proximity-based split in Eq.~\ref{eq:prox_split} when there exists a pivot $p_+$ such that $\beta_i = d(\alpha_i^\intercal x, \alpha_i^\intercal p_+)$. 
Thus, Eq.~\ref{eq:prox_split} can be seen as a generalization for all the binary splits discussed in Section~\ref{sec:background}.

Now, consider training a \textsc{udt}, and a \textsc{hdt}, on the same dataset under identical greedy strategy, impurity measure, and hyper-parameters, such as maximum depth. 
Suppose that at iteration $i$, \textsc{udt} selects the best feature-based splitting function according to the information gain $\Delta_i(\alpha_i, \beta_i)$. 
By definition, there does not exist another pair $(\alpha_i', \beta_i') \neq (\alpha_i, \beta_i)$ such that $\Delta_i(\alpha_i', \beta_i') > \Delta_i(\alpha_i, \beta_i)$, otherwise it would have been chosen.

However, while \textsc{udt} only evaluates axis-parallel splits, i.e., with $\sum_{j=1}^m \alpha_{i,j} = 1$, \textsc{hdt} explores a richer set of possibilities, including oblique splits, selective feature splits, distance-based splits, proximity-based splits, and their respective variants. 
Thus, for each iteration $i$, given the same dataset $X$, \textsc{hdt} can potentially identify a pair $(\alpha_i', \beta_i')$ such that $\Delta_i(\alpha_i', \beta_i') \geq \Delta_i(\alpha_i, \beta_i)$, thanks to the broader expressiveness of its splitting functions.
This behavior, repeated across multiple levels of the tree, ensures that the hybrid tree can never have lower information gain at any node compared to any homogeneous counterpart. 
Indeed, the same argument holds not only w.r.t. \textsc{udt}, but also when comparing trees based on any fixed, homogeneous split type.
A higher information gain at a split leads to purer child nodes more rapidly, implying improved generalization and potentially more imbalanced but accurate tree structures. 
Therefore, between two shallow trees, one with homogeneous splits and one with heterogeneous splits, trained under the same maximum depth imposed constraint, for example, for interpretability, the hybrid tree has the potential to be more accurate.
\textcolor{black}{While \textsc{hdt} guarantees higher information gain per split, this does not necessarily translate into higher predictive accuracy, as smaller-gain splits can sometimes yield better-performing trees. 
Also, the added flexibility of \textsc{hdt} may increase the risk of overfitting, though a shallow \textsc{hdt} can overfit less than a deeper \textsc{udt}. 
Still, these considerations provide a solid rationale for the design of our proposal.}

\section{Experiments}
\label{sec:experiments}
We present here an empirical evaluation of \textsc{HybridDecisionTree} (\textsc{hdt}) and \textsc{SelectivePivotTree} (\textsc{spt}). 
First, we assess the performance of our proposals against a broad range of competitive baselines, including standard 
homogeneous feature-based and case-based 
\textsc{dt} models, and ensemble-based tree learners, across a diverse collection of datasets.
The implementation choices adopted are motivated by an ablation study on \textsc{hdt} and a sensitivity analysis on \textsc{spt}.

\smallskip
\textbf{Experimental Setting.}
We experimented on 25 datasets\footnote{Data and sources are available at \url{https://github.com/fismimosa/Hybrid-Decision-Tree}. The authors implement all the tree-based models used in this work. Ensemble learners are available at: \url{https://scikit-learn.org/stable/}, \url{https://xgboost.readthedocs.io/}, \url{https://interpret.ml/docs/ebm}. Server specifications: 2xAMD EPYC 9754 128-core.}.
with distinct characteristics, summarized in Table~\ref{tab:datasets_info}. 
For each dataset, we adopt an 80-20 stratified train-test split. 
For model selection, we employ a stratified $5$-fold cross-validation on the training set to select the best hyperparameters based on macro F1-score, and report final performance on the held-out test set.
Since some models do not support non-numerical data, we apply one-hot encoding to categorical features.
We evaluate classification performance using the macro F1-score and measure model complexity by the number of nodes, favoring more interpretable trees with more compact structures.

We describe here the types of splitting functions $\Gamma$ analyzed:
\begin{itemize}[noitemsep,topsep=0pt]
    \item \textsc{ud}: Univariate, axis-aligned feature-based splits, implemented following the standard \textsc{CART} strategy~\cite{breiman1984classification}.
    
    \item \textsc{od}: Oblique linear splits, based on the HHCART algorithm~\cite{Wickramarachchi2016HHCART}, constrained to at most two non-zero feature coefficients to preserve interpretability.
    
    \item \textsc{p}: Distance-based splits as proposed in \textsc{PivotTree}~\cite{cascione2024pivotree}, which route instances according to their distance to a single pivot, using the condition described in Eq.~\ref{eq:dist_split}.
    
    \item \textsc{pp}: A variant of \textsc{p} split that selects the best pair of candidate pivots from opposing classes, denoted $p_{-}$ and $p_{+}$. Instances are routed based on their relative proximity to these two pivots, using the condition described in Eq.~\ref{eq:prox_split}.

    \item \textsc{sp}: A selective pivot split using a distance-based strategy, implemented as described in Alg.~\ref{alg:selective-split} with $\pi = \mathit{False}$.
    
    \item \textsc{spp}: A selective proximity pivot split employing a proximity-based strategy, with $\pi = \mathit{True}$.

\end{itemize}
We denote homogeneous trees constructed using a single split type by appending the suffix \textsc{t} to the split type, e.g., \textsc{udt} for univariate feature-based splits, \textsc{ppt} for proximity-based splits, extending this convention to other cases similarly.
To conduct the ablation study, we assess \textsc{hdt} variants by modifying the set of split functions $\Gamma$, using the notation $\textsc{hdt}^{-}_{\mathit{splits}}$ to indicate which split types are excluded: e.g., $\textsc{hdt}^{-}_{\textsc{p}}$ excludes \textsc{p} splits; $\textsc{hdt}^{-}_{\textsc{sp},\textsc{spp}}$ excludes both selective types.
In particular, we consider \textsc{hdt} including all splitting functions described above, along with $\textsc{hdt}^{-}_{\textsc{sp}}$ and $\textsc{hdt}^{-}_{\textsc{spp}}$. 
This selection is motivated by the ablation study on \textsc{hdt} reported in the following.

As optimal configuration, we adopt $v = 0.1$, $k = 100$, \textsc{fie} for \textsc{SelectableFeatures}, and uses the clustering-based (\textsc{c}) implementation for \textsc{CandidatePivots} in \textsc{hdt}, while employing the random (\textsc{r}) implementation for \textsc{spt} and \textsc{sppt}.
We refer the readers to the paper repository for precise configurations used for other models.
These settings are motivated by experiments presented at the end of this section.
Given that case-based splits are more sensitive to data preprocessing than feature-based ones, we also consider each case-based split under three preprocessing strategies: none, \textit{MinMax Scaling}, and \textit{Standard Scaling}. 
Both scalers are fit on the training set and applied only to the instances evaluated by case-based splits during tree construction. 
The variant yielding the highest impurity gain, including those without scaling, is selected.
All models, homogeneous and hybrid, are evaluated using $\mathit{max\_depth} \in [1, 10] \cup \{\infty\}$. 
When $\mathit{max\_depth} = \infty$, tree growth halts when no further gain is possible. 
For case-based splits, we use the Euclidean distance as the metric $d$.
All statistical significance tests are performed using the Nemenyi test with $\alpha = 0.1$, consistently across all experiments~\cite{demvsar2006statistical}.

\begin{figure}[t]
    \centering
    \includegraphics[trim={0 87mm 0 0mm},clip,width=.9\linewidth]{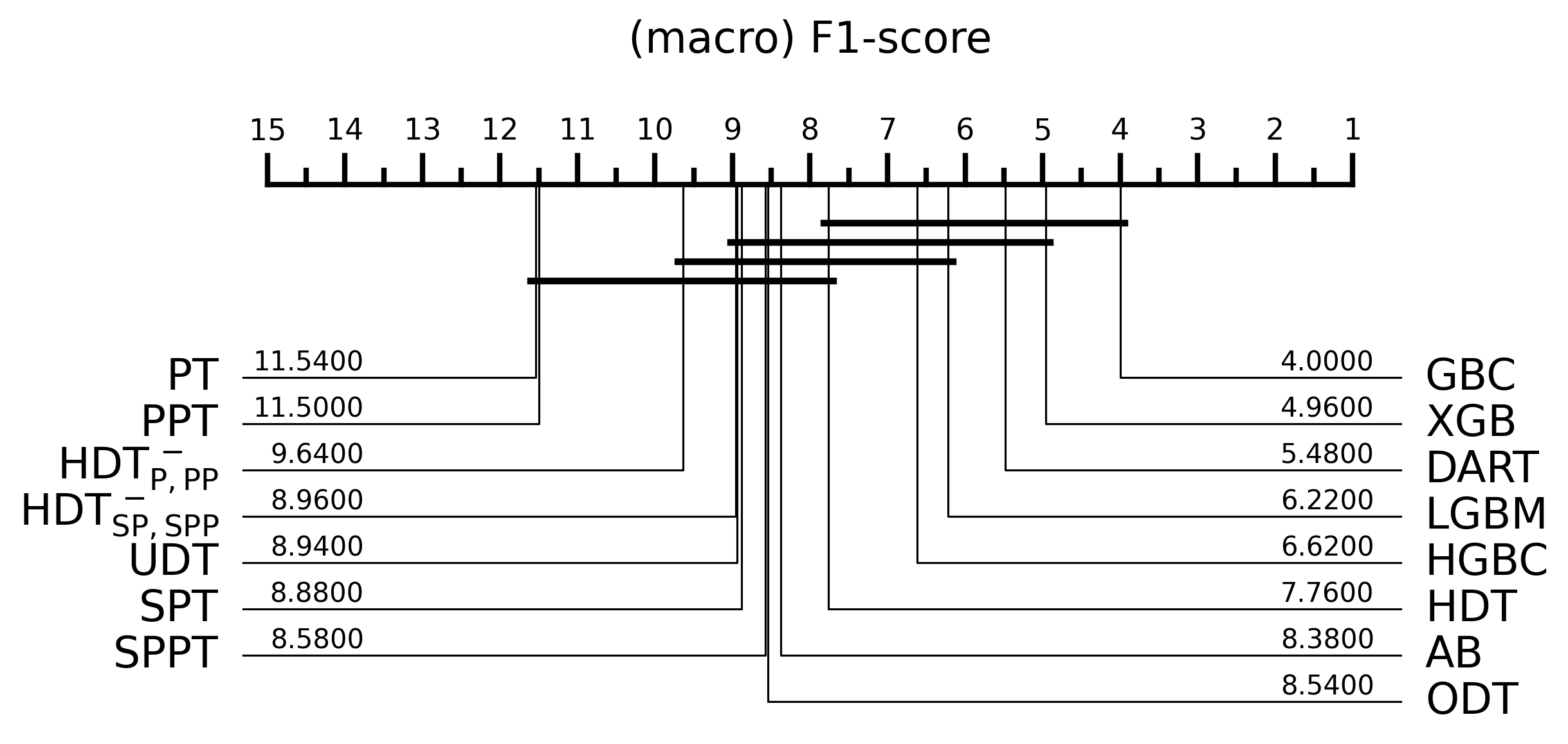}
    \includegraphics[trim={0 60mm 0 15mm},clip,width=.9\linewidth]{z_img_3Final_cd_F1.png}
    \includegraphics[trim={0 3,5mm 0 45mm},clip,width=.9\linewidth]{z_img_3Final_cd_F1.png}
    \caption{Models ranking by F1-score. Statistically indistinguishable models are connected; best ranks on the right.}
\label{fig:z_img_3Final_cd_F1}
\end{figure}

\smallskip
\textbf{Comprehensive Model Assessment.}
We evaluate the performance of \textsc{hdt} and \textsc{spt} against homogeneous \textsc{dt}s built with all previously described splitting functions.
Furthermore, we compare against ensemble models based on homogeneous \textsc{udt}s to assess whether a standalone hybrid approach can match the performance of standard tree ensembles.
We consider the following ensemble~\cite{Ibomoiye2022surveyens}: Random Forest (\textsc{rf}), AdaBoost (\textsc{ab}), Extremely Randomized Trees (\textsc{erts}), HistGradientBoostingClassifier (\textsc{hgbc}), GradientBoostingClassifier (\textsc{gbc}), eXtreme Gradient Boosting (\textsc{xgb}), Dropout and Additive Regression Trees (\textsc{dart}), Explainable Boosting Classifier (\textsc{ebc})\footnote{We additionally tested \textsc{gosdt}~\cite{Lin2020Generalized}, but, due to the high computational complexity, were able to obtain results only on the top 10 smaller datasets, which was insufficient for comparison with other methods and for conducting a solid statistical test. 
}.
We first perform a stratified $5$-fold cross-validation on the \texttt{small} subset of the datasets listed in Table~\ref{tab:datasets_info}. 
This step aims to identify the best overall hyperparameter configuration for each model across different datasets. 
Specifically, in line with~\cite{Middlehurst2024bakeoff}, for each model, we select the configuration that performs best on the majority of datasets during validation.
The selected configuration is used in the assessment phase, where models are trained on the full training set and evaluated on the test set across all datasets in Table~\ref{tab:datasets_info}.

\begin{figure}
    \centering
    \includegraphics[trim={0 2mm 0 2mm},clip,width=0.8\linewidth]{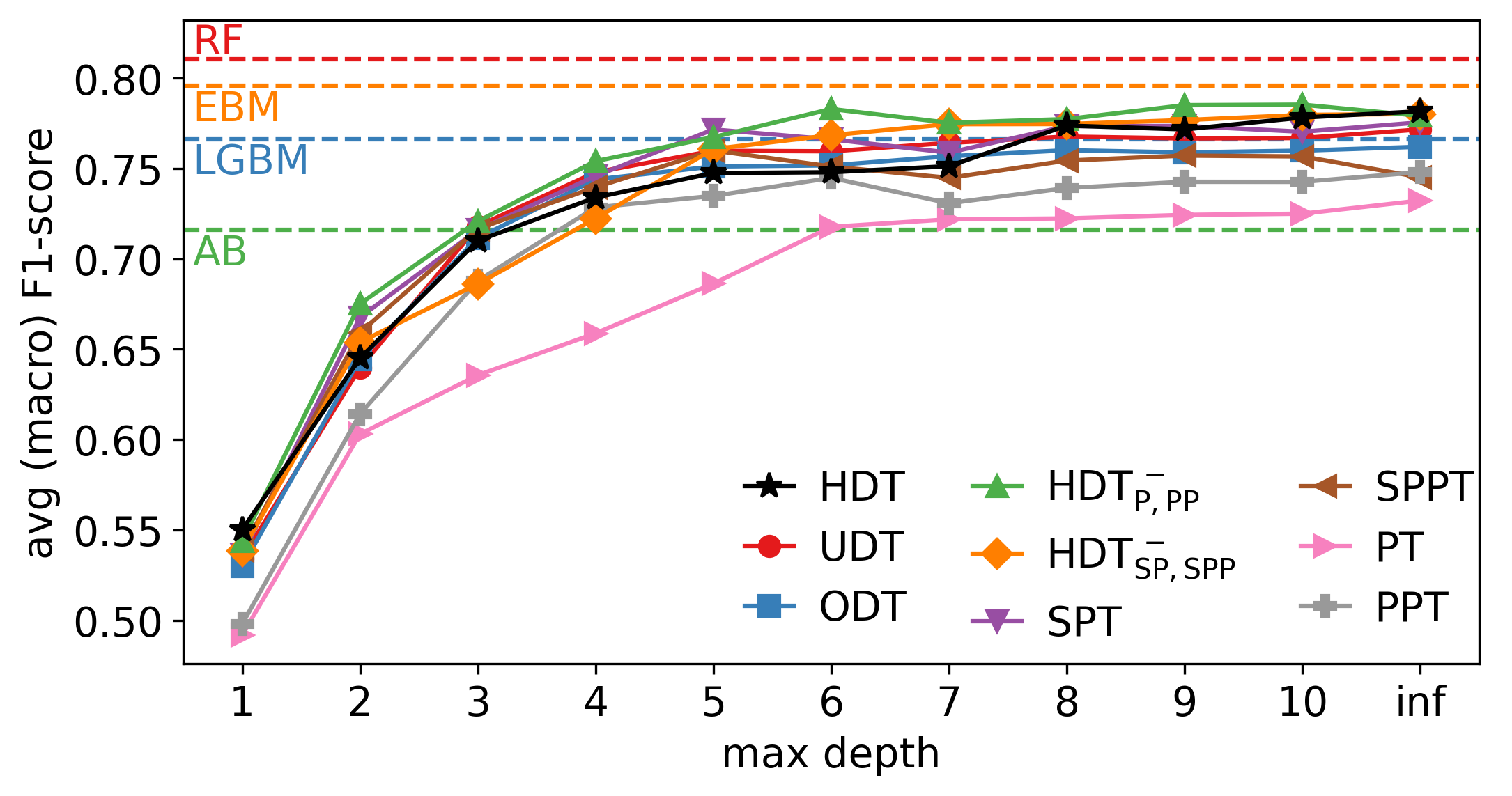}
    \caption{Average (macro) F1-score while varying  maximum tree depths.
    Ensemble models are marked with horizontal dotted lines.}
    \label{fig:z_img_3Final_scatter_F1_vs_depth}
\end{figure}

Figure~\ref{fig:z_img_3Final_cd_F1} shows a CD plot comparing models in terms of macro F1-score\footnote{Detailed dataset-specific results are available in the repository.}, where top-performing models appear on the right, and statistically indistinguishable ones are connected.
Results highlight that the considered \textsc{hdt} variants achieve higher performance compared to homogeneous trees. 
Moreover, they are competitive with ensemble-based methods, yielding statistically indistinguishable results from these models according to our evaluation.
We complement the above results with an analysis of model complexity. 
In Figure~\ref{fig:z_img_3Final_scatter_F1_vs_depth}, we illustrate how the macro F1-score varies w.r.t. the maximum tree depth for single tree-based models.
Instead, for ensemble models, the hyperparameters are fixed.
We observe that for shallower trees, \textsc{spt} and \textsc{hdt} variants using also selective splits achieve competitive performance. 
Notably, $\textsc{hdt}^{-}_{\textsc{p},\textsc{pp}}$ that excludes non-selective case-based splits outperforms some ensemble models even at moderate depths like 5 or 6, and consistently surpasses all competing standalone tree models.

\begin{figure}[t]
    \centering
    \includegraphics[trim={0 75mm 0 0mm},clip,width=.9\linewidth]{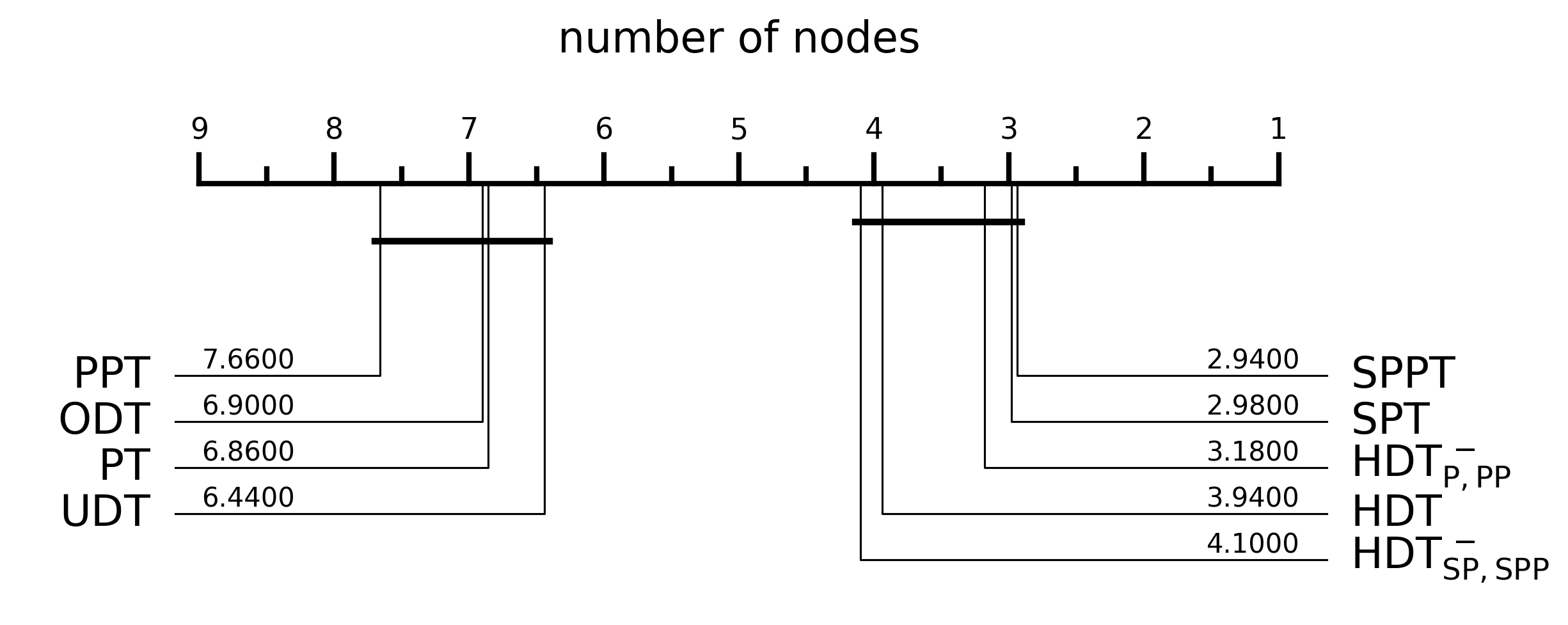}
    \includegraphics[trim={0 45mm 0 15mm},clip,width=.9\linewidth]{z_img_3Final_cd_F1_nodi_all_hyper.png}
    \includegraphics[trim={0 5mm 0 45mm},clip,width=.9\linewidth]{z_img_3Final_cd_F1_nodi_all_hyper.png}
    \caption{Models ranking by complexity as number of nodes. Statistically indistinguishable models are connected; best ranks on the right.}
\label{fig:z_img_3Final_cd_F1_nodi_all_hyper}
\end{figure}

In Figure~\ref{fig:z_img_3Final_cd_F1_nodi_all_hyper}, we examine the interpretability of \textsc{hdt} and \textsc{spt} variants in terms of complexity by comparing the number of nodes in their best-performing configurations to those of homogeneous trees. 
Tree sizes may vary due to early stopping based on impurity gain thresholds.
Both \textsc{hdt} and \textsc{spt} variants consistently rank among the most compact models.
\textsc{hdt} and \textsc{spt} variants consistently yield more compact trees. 
Notably, \textsc{sppt}, \textsc{spt}, \textsc{hdt} and its variants are statistically distinct from homogeneous baselines, showing strong performance with fewer nodes. While the four selective pivot split trees are top performers, even $\textsc{hdt}^-_{SP, SPP}$, which excludes these splits, achieves a statistically significant reduction in node count.

Finally, we report that \textsc{hdt} and \textsc{spt} variants are slower than their homogeneous counterparts in both training and inference\footnote{Detailed training and inference runtimes are provided in the repository.}.
Our implementation reports average training times 390s for \textsc{hdt}, 170s for \textsc{spt}, 435s for \textsc{sppt}, and just 0.15s for \textsc{udt}. 
Inference takes 0.32s, 0.52s, and 0.91s for \textsc{hdt}, \textsc{spt}, and \textsc{sppt}, versus 0.0001s for \textsc{udt}.
This overhead arises from evaluating multiple split types in \textsc{hdt}, and from feature selection and exhaustive search in \textsc{spt}.
Future work will explore strategies like bagging to limit the number of features evaluated per node and reduce computational cost.

\begin{figure}[t]
    \centering
    \includegraphics[trim={0 90mm 0 0mm},clip,width=.9\linewidth]{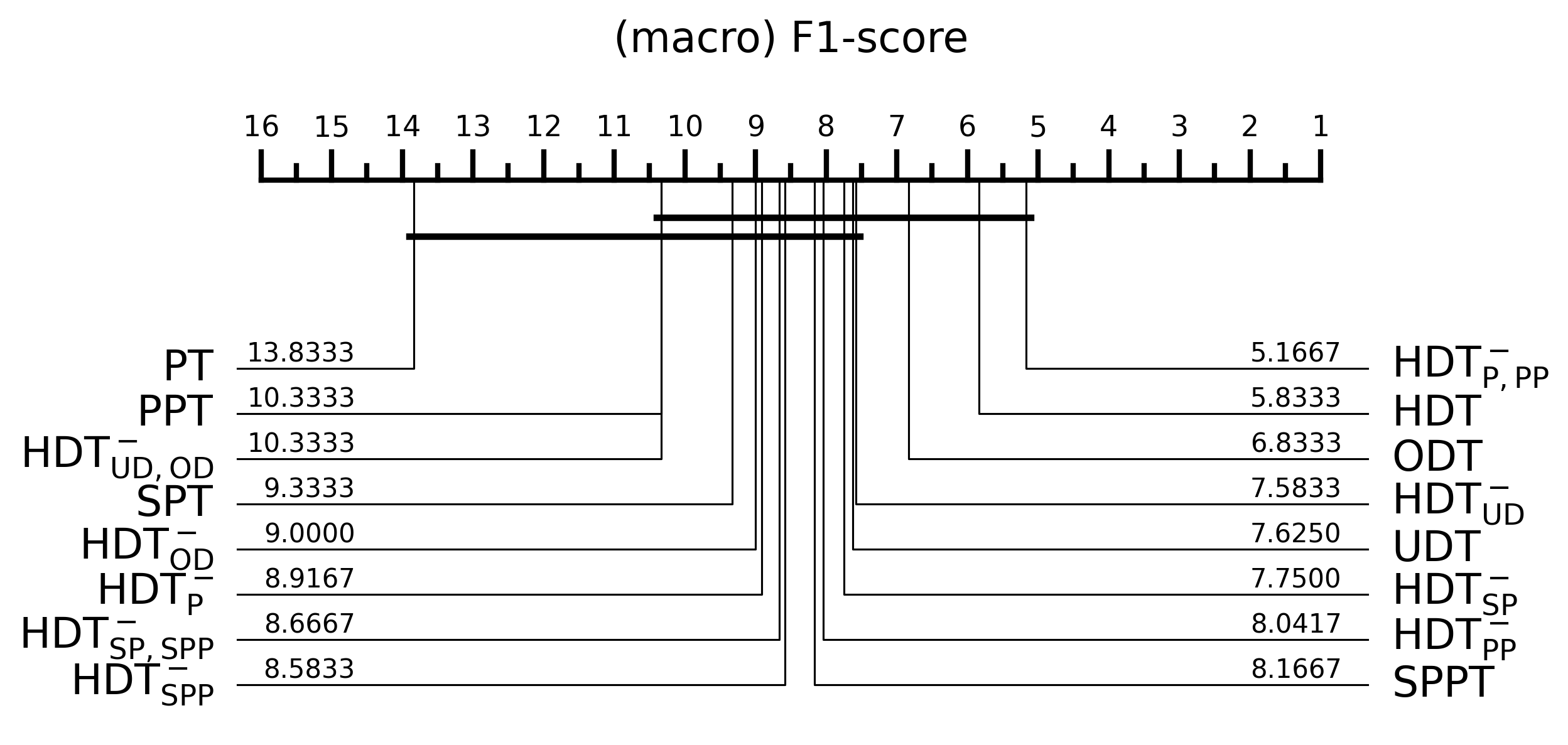}
    \includegraphics[trim={0 68mm 0 15mm},clip,width=.9\linewidth]{z_img_2Ablation_depth4_F1.png}
    \includegraphics[trim={0 5mm 0 43mm},clip,width=.9\linewidth]{z_img_2Ablation_depth4_F1.png}
    \caption{\textsc{hdt} variants ranking by F1-score. Statistically indistinguishable models are connected; best ranks on the right.
    }
\label{fig:ab_cdplot_f1}
\end{figure}

\begin{figure}[t]
    \centering
    \includegraphics[trim={0 90mm 0 0mm},clip,width=.9\linewidth]{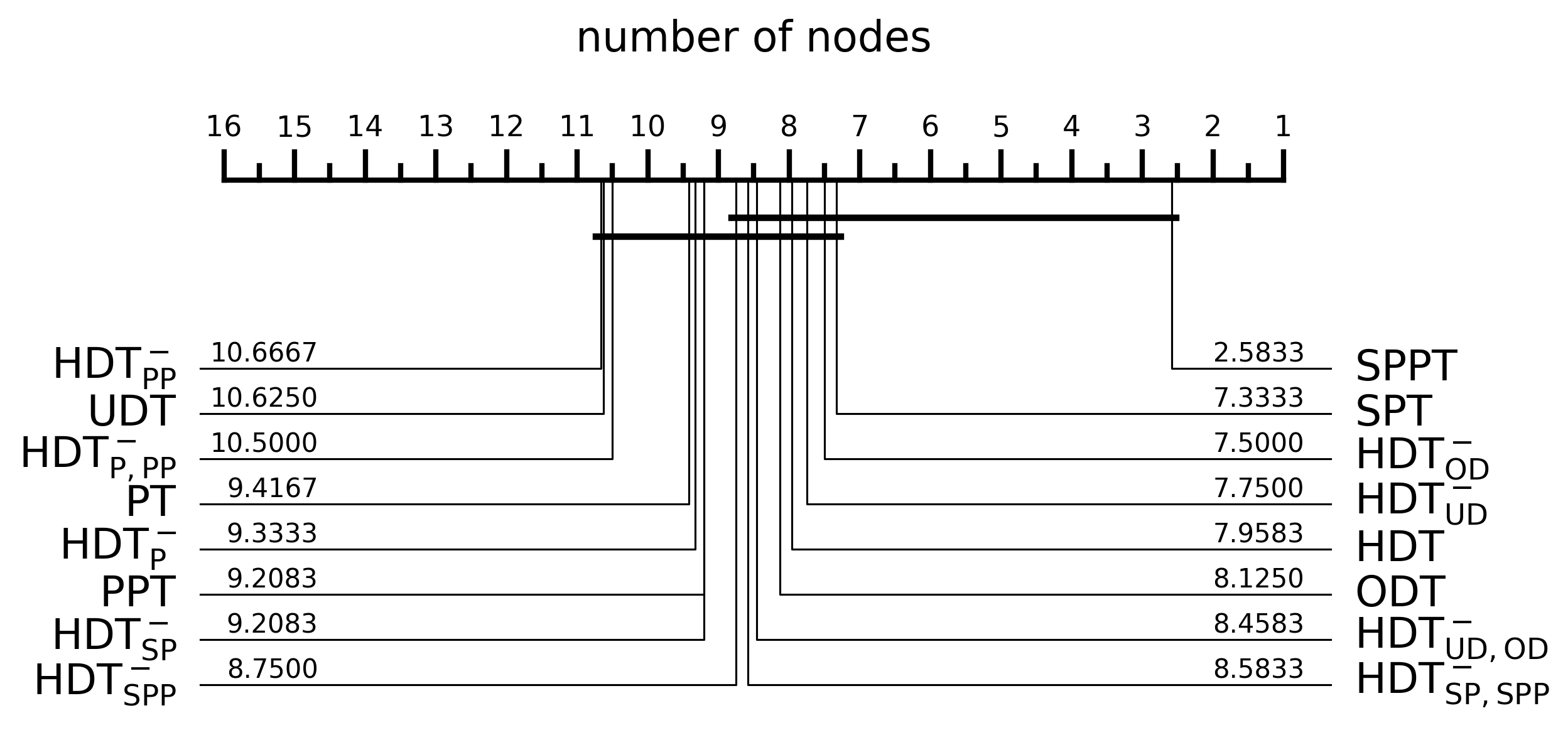}
    \includegraphics[trim={0 67,8mm 0 15mm},clip,width=.9\linewidth]{z_img_2Ablation_depth4_nodi.png}
    \includegraphics[trim={0 4mm 0 45mm},clip,width=.9\linewidth]{z_img_2Ablation_depth4_nodi.png}
    \caption{\textsc{hdt} variants ranking by complexity as number of nodes. Statistically indistinguishable models are connected; best ranks on the right.}
\label{fig:ab_cdplot_nodes}
\end{figure}


\smallskip
\textbf{Ablation Study.} 
This section analyzes how removing specific split types from the candidate pool $\Gamma$ in Algorithm~\ref{alg:hybrid-tree} influences the predictive performance and complexity of \textsc{hdt}.
For completeness, we also include in the analysis \textsc{spt}s and \textsc{sppt}s using the same configurations adopted in the comprehensive evaluation.
We conduct the ablation study on the \texttt{small} datasets listed in Table~\ref{tab:datasets_info}, specifically those containing at most 5k instances and no more than 10 features.
In this setting, according with~\cite{bertsimas2017optimal}, we fix $\mathit{max\_depth} = 4$. 

\begin{figure}[t]
    \centering
    \includegraphics[trim={0 4mm 0 6mm},clip,width=0.9\linewidth]{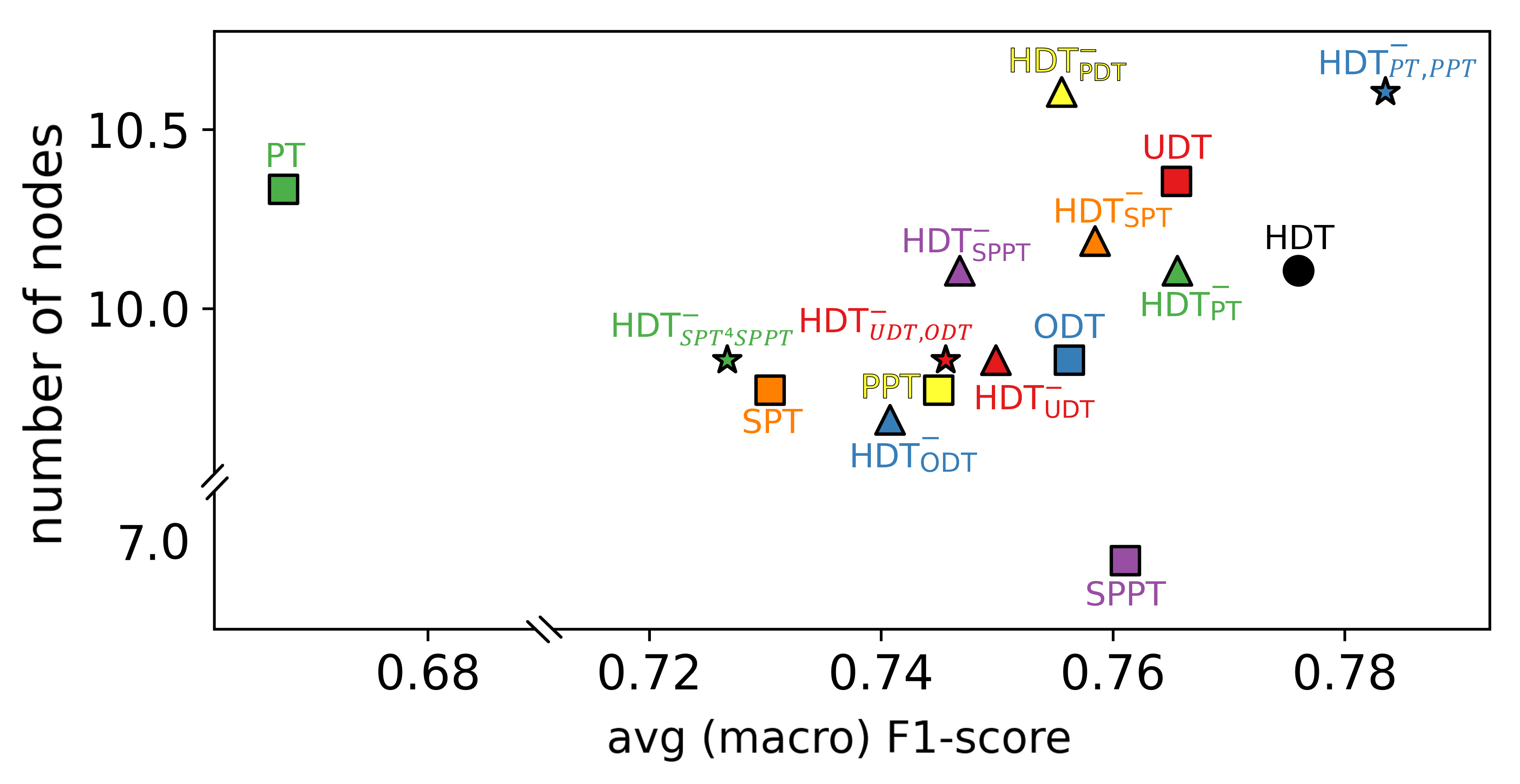}
    \caption{F1-score vs number of nodes for \texttt{small} datasets.}
    \label{fig:scatter_ablation}
\end{figure}

Figures~\ref{fig:ab_cdplot_f1} and~\ref{fig:ab_cdplot_nodes} present CD plots comparing \textsc{hdt} variants in terms of macro F1-score and number of nodes.
While Figure~\ref{fig:ab_cdplot_f1} shows that \textsc{hdt} models are statistically on par with homogeneous baselines like \textsc{udt} and \textsc{odt}, they consistently deliver better overall performance.
We highlight how variants lacking \textsc{sp} and \textsc{spp} strategies show reduced performance and less tree compactness, emphasizing the value of selective strategies.
The top-performing \textsc{hdt} model is \textsc{hdt}$^{-}_{\textsc{p},\textsc{pp}}$, followed by \textsc{hdt}.
In terms of complexity, Figure~\ref{fig:ab_cdplot_nodes} shows that \textsc{hdt} variants, along with \textsc{spt}s and \textsc{sppt}s, produce considerably shallower trees than homogeneous models like \textsc{udt}, \textsc{pt}, and \textsc{odt}.
Figure~\ref{fig:scatter_ablation} shows the trade-off between predictive performance and model size across \textsc{hdt} variants.
The results confirm previous CD plot insights, highlighting the compactness and strong performance of \textsc{sppt} models.
Overall, \textsc{hdt} variants offer the best balance between accuracy and complexity compared to homogeneous and hybrid alternatives.

\begin{figure}[t]
    \centering
    \includegraphics[trim={0 2mm 0 3mm},clip,width=0.9\linewidth]{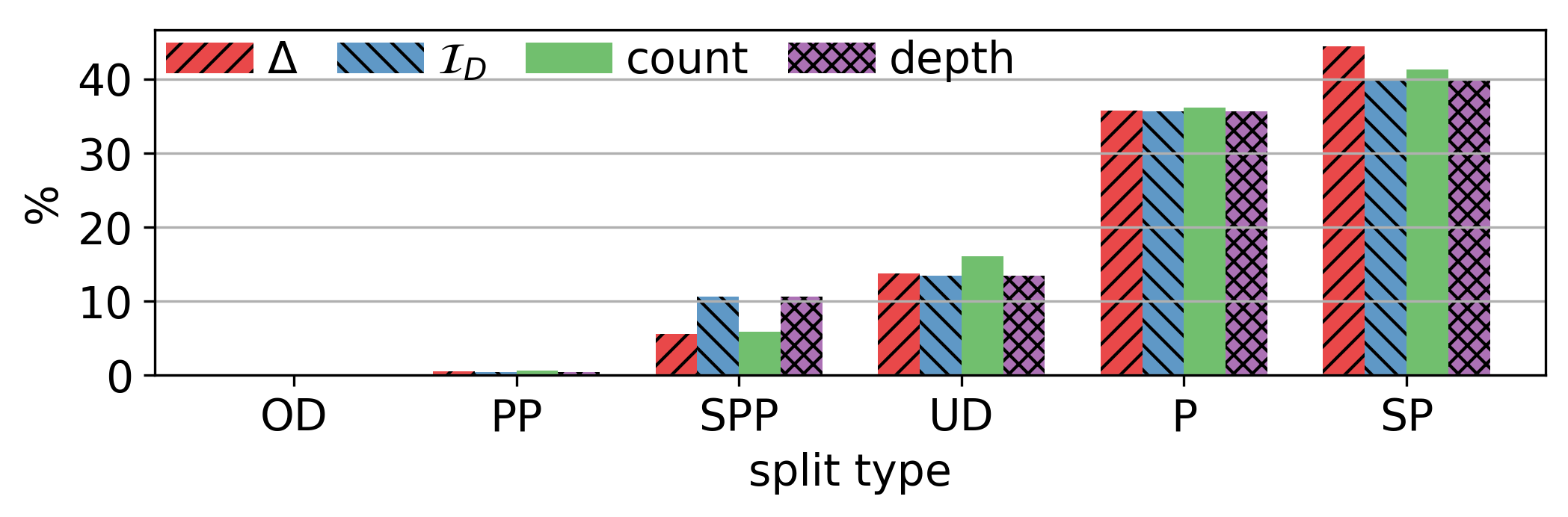}
    \caption{Relevance of different splitting strategies in hybrid trees evaluated by frequency, average depth, impurity $\mathcal{I}_D$, and information gain $\Delta$.}
    \label{fig:ablation_barplot}
\end{figure}

Figure~\ref{fig:ablation_barplot} provides an analysis of the impact of each splitting function within \textsc{hdt} models. 
We assess their contribution based on four criteria: frequency of use (count), inverse average depth of application (depth), Gini impurity reduction ($\mathcal{I}_D$), and (iv) information gain ($\Delta$) as defined in Eq.~\ref{eq:information_gain}.
The results reveal that the \textsc{sp} split is the most frequently selected and yields the highest information gain, highlighting its effectiveness in enhancing node-level decisions. 
The \textsc{p} split follows as the second most influential, with the standard \textsc{ud} split ranking third.
As a complement to the theoretical discussion presented in Section~\ref{sec:method}, we provide an empirical verification of a key property: at the root level, \textsc{hdt} trees consistently achieve a higher information gain compared to all homogeneous competitors. 
This supports the theoretical intuition that hybrid splitting strategies lead to more informative data partition.
Furthermore, we empirically compared the information gain at the root node between \textsc{hdt} and \textsc{udt}.
On average, \textsc{hdt} achieved an information gain of $0.20$ compared to $0.18$ for \textsc{udt} on \texttt{small} datasets, and $0.23$ versus $0.21$ on \texttt{large} datasets, highlighting the superior split quality of hybrid models across dataset scales.

Based on the results of the ablation study, we selected the following hybrid tree configurations for the comprehensive model assessment: the full \textsc{hdt} and $\textsc{hdt}^{-}_{\textsc{p},\textsc{pp}}$, identified as the best-performing models, and $\textsc{hdt}^{-}_{\textsc{sp},\textsc{spp}}$ to further investigate the impact of the proposed selective splitting strategies.

\smallskip
\textbf{Sensitivity Analysis.}
We conduct here a focused evaluation of selective splits to identify configurations that best balance predictive performance and training efficiency, using the \texttt{small} datasets with $\mathit{max\_depth} = 3$. 
For each hyperparameter, we analyze its impact by varying its value, keeping others fixed to a default setting available on the repository.
The binary flag $\pi$ is included to choose between distance-based (\textsc{sp}) and proximity-based (\textsc{spp}) splits.
We assess four preprocessing options: \textit{No} scaling, \textit{MinMax}, \textit{Standard}, and \textit{MaxAbs} scaling.
For instance selection via \textsc{CandidateInstances}, we test $v {\in} \{0.1, 0.5, 1.0\}$, where $v {=} 1.0$ means no filtering, i.e., $X' = X$. 
Also, we consider fixed-size selections with $v/|X| \in \{10, 100, 1000\}$, allowing us to explore values of varying magnitudes.
For \textsc{SelectableFeatures}, we compare: \textsc{fc} with $h \in \{1, \dots, 5\}$, \textsc{gc}, and feature-importance-based \textsc{fie} and \textsc{fid}, using Random Forests\footnote{We chose Random Forests due to their speed and reliability, especially compared to linear models like logistic regression. To reduce variance, we used a forest with 1000 trees. All other hyperparameters were kept at their default values as defined by the underlying library.}.
Finally, for \textsc{CandidatePivots}, we evaluate \textsc{a} (all), \textsc{r} (random), and \textsc{c} (clustering), with $k \in \{10, 100\}$ for the latter two\footnote{We selected these two values to span different orders of magnitude while ensuring reasonable computational times.}.

\begin{figure*}
    \centering    \includegraphics[width=.85\textwidth]{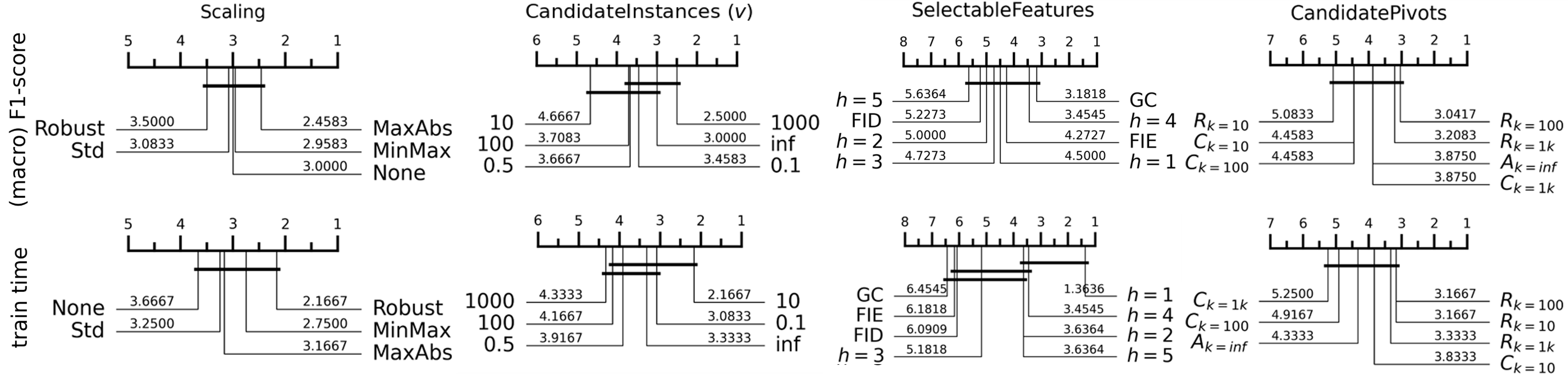}
    \caption{Selective pivot split configurations ranked by F1-score and training time. Statistically tied setups are connected, best ranks to the right.}
    \label{fig:cdplot_COMPLETE_SENSITIVITY}
\end{figure*}

Figure~\ref{fig:cdplot_COMPLETE_SENSITIVITY} compares selective split models by F1-score and training time across configurations.
No preprocessing method showed clear superiority, so both \textit{MinMax}, \textit{Standard}, and \textit{No} scaling were retained, the latter because, despite its weaker theoretical grounding for distance-based calculations, it empirically provided a notable performance boost and can enhance interpretability.
Sampling 10\% of training data ($v = 0.1$) for \textsc{CandidateInstances} proved optimal balancing between accuracy and speed.
All \textsc{SelectableFeatures} methods performed comparably, supporting the choice of \textsc{fie} for its efficiency.
For \textsc{CandidatePivots}, selecting 100 pivots via \textsc{r} and \textsc{c} offered the best trade-off, also yielding slightly smaller trees than with $k=10$.
These findings validate the selected \textsc{spt} setup as both effective and efficient.

\section{Conclusion}
\label{sec:conclusion}
We have introduced \textsc{HybridDecisionTree}, a tree induction strategy combining feature-based and case-based splits, and extended it with the \textsc{SelectivePivotTree}, which performs distance-based splits over a selected feature subset.
Experiments on tabular datasets show that our hybrid models outperform standard trees in both accuracy and interpretability, particularly at shallow depths, and that selective splitting plays a crucial role in reaching such performance.
Remarkably, the increased expressive power of \textsc{HybridDecisionTree} allows for shallow trees to rival or surpass traditional ensemble methods.
For future work, we plan to extend these models to evolutionary~\cite{Kretowski2004evol} and generative~\cite{guidotti2024gentree} learning frameworks, integrate interpretable learned distance metrics~\cite{Fedele2025Interpretable,yang2006distance}, and compare them against stronger baselines such as optimally~\cite{bertsimas2017optimal} and generative~\cite{guidotti2024gentree} learned trees.
We also aim to develop hybrid ensembles combining diverse splits and improve training efficiency through bagging and feature reduction techniques.
We also consider validating our approach in a decision-making setting with human subjects to assess its practical value.

\section{Conclusion}
\label{sec:conclusion}
We have introduced \textsc{HybridDecisionTree}, a tree induction strategy combining feature-based and case-based splits, and extended it with the \textsc{SelectivePivotTree}, which performs distance-based splits over a selected feature subset.
Experiments on tabular datasets show that our hybrid models outperform standard trees in both accuracy and interpretability, particularly at shallow depths, and that selective splitting plays a crucial role in reaching such performance.
Remarkably, the increased expressive power of \textsc{HybridDecisionTree} allows for shallow trees to rival or surpass traditional ensemble methods.
For future work, we plan to extend these models to evolutionary~\cite{Kretowski2004evol} and generative~\cite{guidotti2024gentree} learning frameworks, integrate interpretable learned distance metrics~\cite{Fedele2025Interpretable,yang2006distance}, and compare them against stronger baselines such as optimally~\cite{bertsimas2017optimal} and generative~\cite{guidotti2024gentree} learned trees.
We also aim to develop hybrid ensembles combining diverse splits and improve training efficiency through bagging and feature reduction techniques.
We also consider validating our approach in a decision-making setting with human subjects to assess its practical value.

\section*{Acknowledgments}
This work has been partially supported by the Italian Project Fondo Italiano per la Scienza FIS00001966 ``MIMOSA'', by the European Community Horizon~2020 programme under the funding schemes, 
G.A. 101070212 ``FINDHR'', G.A. 101120763 ``TANGO'', by the European Commission under the NextGeneration EU programme – National Recovery and Resilience Plan (Piano Nazionale di Ripresa e Resilienza, PNRR) Project: ``SoBigData.it – Strengthening the Italian RI for Social Mining and Big Data Analytics'' – Prot. IR0000013 –  Av. n. 3264 del 28/12/2021, and M4C2 - Investimento 1.3, Partenariato Esteso PE00000013 - ``FAIR'' - Future Artificial Intelligence Research'' - Spoke 1 ``Human-centered AI''.

\bibliographystyle{IEEEtran}
\bibliography{IEEEabrv,biblio_short}

\end{document}